\documentclass[runningheads]{llncs}
\usepackage{amsmath}
\usepackage{multirow}
\usepackage[T1]{fontenc}
\usepackage{graphicx,verbatim}
\usepackage{marvosym}
\begin{document}
\title{Unlocking the Power of Medical Tabular Data via Semantic-Aware Multimodal Pre-training}
\titlerunning{Unlocking the Power of Medical Tabular Data}
%
\author{Yingsheng Liu\inst{1,2} \and
Haiming Li\inst{2} \and
Jingmin Zhu\inst{1} \and Jiajun Sun\inst{2} \and
Victoria Mar\inst{2} \and
Monika Janda\inst{3} \and H. Peter Soyer\inst{3} \and
Zongyuan Ge\inst{2} \and
Zhen Yu\inst{2}(\Letter)}
%

\authorrunning{Y.Liu et al.}
%
\institute{Faculty of Information Technology, Monash University, Melbourne, Australia \and
Monash University, Melbourne, Victoria, Australia \and
The University of Queensland, Brisbane, Queensland, Australia\\
\email{Zhen.Yu1@monash.edu}}

\maketitle              
\begin{abstract}
While vision-language models dominate medical representation learning, unstructured text lacks the dense, quantitative diagnostic phenotypes inherent in structured clinical tables. However, existing multimodal pre-training methods underutilize this potential due to semantic-agnostic designs that treat tabular inputs as flat vectors and employ unstable continuous regression objectives. To overcome this, we propose a novel semantic-aware framework explicitly modeling the intrinsic two-dimensional structure of tabular data. First, addressing the inter-feature hierarchy of varying diagnostic importance, we introduce Importance-Aware Adaptive Masking to construct a label-free curriculum prioritizing salient features. Second, addressing the intra-feature continuity-discreteness duality, we propose a Soft-Label Discretized Module that replaces unstable numerical regression with stable distribution matching, thereby mathematically preserving ordinal relationships. Extensive experiments across large-scale dermatology (SLICE-3D, HOP) and ophthalmology (EyePACS) datasets establish a new state-of-the-art (SOTA), demonstrating exceptional robustness and cross-domain generalizability.
\noindent\textbf{Code:} \url{https://github.com/Ethan-ysliu/AID}

\keywords{Multimodal  \and Self-supervised Learning \and Image-tabular.}
\end{abstract}

\section{Introduction}

\begin{figure}[t]
  \centering
  \includegraphics[width=0.9\columnwidth]{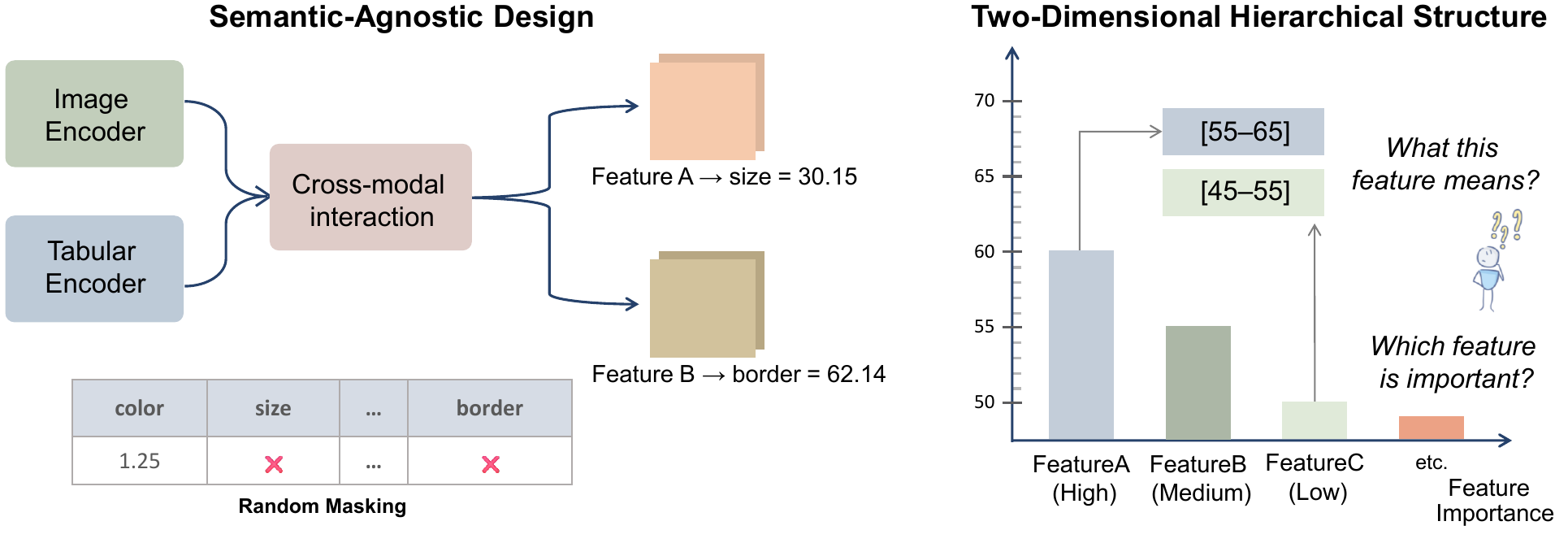}
  \caption{Limitations of Current Semantic-Agnostic Designs and Motivation for the Two-Dimensional Hierarchical Structure. }
  \label{fig:motivation}
\end{figure}


Clinical decision-making is a multimodal process synthesizing visual evidence and structured patient data~\cite{huang2020fusion}. For instance, a radiologist interpreting a chest X-ray benefits from patient smoking history and lab results, while a dermatologist evaluating a skin lesion considers demographic and morphological features, such as age and border irregularity. Although vision-language models have revolutionized representation learning\cite{yan2025make,li2025multi,yan2025derm1m,yan2026vision}, unstructured text often lacks the dense, quantitative diagnostic phenotypes inherent in structured tabular records\cite{hollmann2025accurate}. Large-scale medical databases present a significant opportunity to develop holistic multimodal AI systems utilizing this specific pairing. However, profound label scarcity limits this potential, as obtaining high-quality expert annotations remains an expensive bottleneck\cite{du2024tip}. This reality creates a compelling need for self-supervised learning methods capable of harnessing unlabeled image-tabular data to acquire robust representations for downstream clinical tasks.


Despite this urgent need, existing multimodal pre-training methods are limited by a semantic-agnostic design\cite{du2024tip,hager2023best}. As in Fig.~\ref{fig:motivation}, SOTA approaches typically treat tabular inputs as flat, unstructured vectors and employ semantically naive pretext tasks. First, uniform masking strategies process all clinical features equally, ignoring inherent disparities in diagnostic importance. Second, and more critically, reconstructing masked values via direct continuous numerical regression (typically mean squared error) creates a severe optimization bottleneck. Specifically, forcing a model to regress exact continuous values from nearly identical visual patterns such as distinguishing visually indistinguishable asymmetry scores of $0.61$ and $0.65$ that conceptually map to the shared semantic category ``moderate asymmetry'' creates an ill-posed objective. Such a noisy learning formulation encourages overfitting and prevents the network from capturing robust, high-level semantic links between visual evidence and clinical concepts.


To unlock the full potential of multimodal medical representation learning, we must move beyond this semantic-agnostic paradigm and explicitly model the intrinsic properties of structured clinical data. We posit that medical tabular data possesses a unique two-dimensional hierarchical structure. At the inter-feature level, clinical attributes exhibit varying degrees of diagnostic importance; for example, lesion border asymmetry carries far more clinical weight than general demographic information. At the intra-feature level, values within a single clinical feature demonstrate a continuity-discreteness duality. While maintaining exact continuous precision is necessary for calculating global cross-modal alignment, local feature reconstruction reflects discrete medical semantic intervals. In clinical practice, precise numerical differences often map to shared diagnostic concepts, such as identifying a measurement simply as abnormally large.


To bridge this semantic gap, we propose Adaptive Importance-guided Discretized reconstruction (AID), a semantic-aware framework modeling these two structural dimensions. To address the inter-feature hierarchy, we introduce a label-free importance-aware adaptive masking strategy. Leveraging a frozen meta-learning prior to estimate data-driven significance without accessing downstream targets, this module constructs a label leakage-free curriculum prioritizing diagnostically salient features. To address the intra-feature duality, an end-to-end soft-label discretized module replaces unstable numerical regression with a stable distribution matching objective. Crucially, by utilizing a triangle kernel to allocate probability mass across adjacent semantic intervals, the soft-labeling mechanism translates exact values into probabilities. This operation mathematically preserves the inherent ordinal relationships of clinical measurements, overcoming the limitations of both direct regression and naive hard-label classification.


While early methods established strong baselines for processing tabular data, integrating them into multimodal frameworks often relies on rudimentary fusion techniques. Recent multimodal self-supervised pre-training approaches achieve strong performance by combining contrastive learning with masked feature reconstruction\cite{du2024tip,devlin2019bert,lee2020biobert}. However, these methods uniformly adopt semantic-agnostic view, treating all clinical features equally and relying on continuous value regression. To the best of our knowledge, no existing pre-training objective explicitly models diagnostic feature importance and continuity-discreteness duality of medical tabular data. Our work fills this critical void by establishing a new semantic-centric paradigm for image-tabular representation learning that translates raw clinical measurements into robust, generalizable representations.

\section{Methodology}
\begin{figure}[t]
  \centering
  \includegraphics[width=\columnwidth]{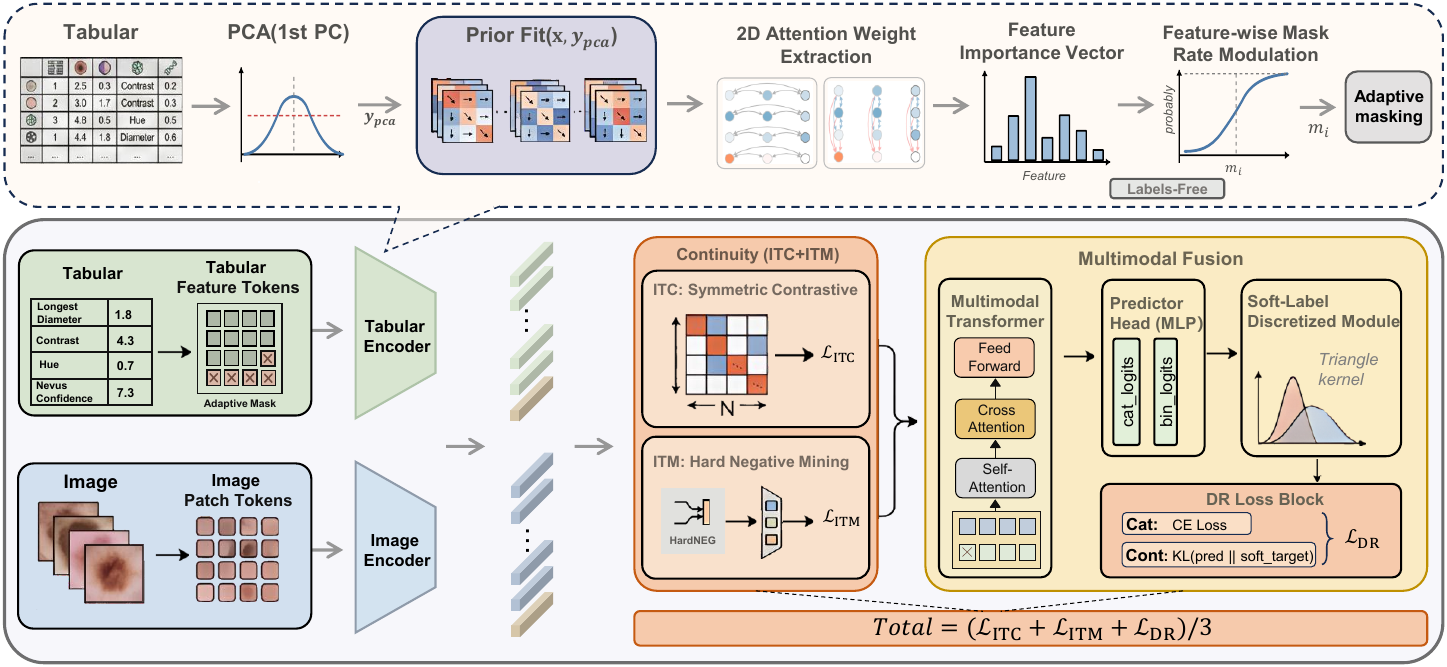}
  \caption{Overall Architecture of the Proposed AID Framework. }
  \label{fig:architecture}
\end{figure}


\subsection{Overall Architecture and the Duality Principle}
As illustrated in Fig.~\ref{fig:architecture}, the AID framework comprises a ViT-base \cite{dosovitskiy2020image} for image encoding, a hybrid Transformer for structured clinical features, and a cross-attention multimodal encoder. The AID framework fundamentally models the continuity-discreteness duality of medical tabular data by processing tabular inputs conditionally based on the optimization objective. For cross-modal alignment and matching, the network utilizes exact continuous numerical values to preserve precise clinical measurement scales. Conversely, for discretized reconstruction, an end-to-end discretized module transforms these continuous values into soft probability distributions over discrete semantic intervals. This duality ensures precise global alignment while compelling the model to learn robust, interval-based clinical semantics during local reconstruction.


\subsection{Feature Importance Extraction \& Adaptive Masking}
To construct a label-free importance-aware masking curriculum, AID extracts data-driven feature significance strictly offline prior to pre-training. Unlike tree-based models requiring downstream ground-truth labels, AID derives importance using an unsupervised approach coupled with a frozen meta-learning prior. Specifically, the framework applies principal component analysis to the standardized tabular feature matrix, utilizing the median of the first principal component to generate balanced binary pseudo-labels. These pseudo-labels adapt a frozen tabular foundation model\cite{hollmann2025accurate}. Since TabPFN v2 is pre-trained on large-scale generative tabular distributions, it provides a universal meta-prior that naturally captures feature dependencies. Fitting this model with pseudo-labels enables its internal self-attention mechanisms to act as a robust inductive bias for dataset-specific feature interactions, mitigating the lack of theoretical guarantees for attention-based feature importance, all without accessing actual diagnostic targets. By registering forward hooks across the attention layers, the system extracts the query, key, and self-attention weight matrices. Aggregating these matrices across all layers, followed by L1 normalization and min-max scaling, produces a normalized global importance vector $s$, representing the relative significance of each clinical feature.

The inter-feature hierarchy of clinical data dictates that specific attributes possess substantially higher diagnostic significance. To embed this inductive bias, the pre-training pipeline utilizes the extracted importance vector $s$ to govern feature corruption. For each continuous feature $j$, AID computes an adaptive masking rate $r_j$:
\begin{equation}
r_j = \min(r_{\text{base}} + \alpha s_j, r_{\max})
\end{equation}
where $r_{\text{base}}$ denotes the base masking probability, $\alpha$ controls the sensitivity to the importance score, and $r_{\max}$ serves as a strict upper bound. This mechanism dynamically establishes a challenging learning curriculum. Highly relevant clinical features receive proportionally higher masking rates, compelling the multimodal encoder to heavily utilize cross-modal visual evidence for reconstructing the masked tabular tokens. To maintain data distribution stability, the masking operation exclusively replaces selected input tokens with a learnable special token embedding, thereby avoiding the representation noise typically introduced by random marginal replacement.


\subsection{Soft-Label Discretized Module}
The intra-feature duality of tabular data necessitates transforming continuous values into stable semantic intervals during reconstruction. Standard hard-label discretization assigns each continuous value to a single bin\cite{bhat2021adabins}, treating semantic intervals independently and destroying the inherent ordinal relationships of clinical measurements. To overcome this limitation, AID introduces an end-to-end soft-label discretized module. Initially, this module establishes $B$ discrete bins per feature using quantile boundaries strictly derived from the training distribution. During the forward pass, for a given feature value $v$ falling into bin $c$ with boundaries $[b_{c-1}, b_c)$, the module computes the intra-bin relative position $p = (v - b_{c-1}) / (b_c - b_{c-1})$. A triangle kernel subsequently formulates a soft-label distribution $q_b$ by allocating probability mass proportionally between the current bin $c$ and its immediate adjacent neighbor ($c-1$ or $c+1$) based on the distance $p$, ensuring $\sum_{b=1}^{B} q_b = 1$. This soft-labeling process dynamically maps precise numerical measurements into smooth probability distributions across adjacent semantic intervals. Distributing mass based on boundary proximity mathematically preserves the ordinality of continuous variables, providing a stable, relation-aware target for cross-modal reconstruction.

\subsection{Multimodal Pre-training}
The comprehensive pre-training objective $\mathcal{L}_{\text{AID}} = (\mathcal{L}_{\text{ITC}} + \mathcal{L}_{\text{ITM}} + \mathcal{L}_{\text{DR}}) / 3$ minimizes a composite loss comprising Image-Tabular Contrastive (ITC), Image-Tabular Matching (ITM), and Discretized Reconstruction (DR) losses. The ITC and ITM objectives align global representations and perform fine-grained binary matching between genuine and mismatched image-tabular pairs using exact continuous values\cite{hager2023best,du2024tip,fu2025unleashing}. Specifically, the ITM objective employs an in-batch hard negative mining strategy, selecting mismatched pairs based on the multimodal similarity scores derived from the ITC computation. Crucially, the adaptive masking process does not degrade the quality of negative mining; the hard negatives are dynamically sampled from the unmasked global representations within the batch, ensuring semantically valid and structurally intact mismatched pairs.

Conversely, the core DR objective reconstructs the masked tabular features utilizing the dual-format output of the predictor head. The DR loss unifies two components: a standard Cross-Entropy loss for the $D_{\text{cat}}$ categorical features, and a Kullback-Leibler (KL) divergence loss for the $D_{\text{con}}$ continuous features. For a batch of $N$ samples, $\mathcal{}{\text{DR}}$ evaluates the divergence between the predicted probabilities (e.g., $p{i,k}$ for continuous bins, $\hat{y}{i,j}$ for categorical classes) and the corresponding targets ($q{i,k}$ from the discretized module, $y_{i,j}$ for true classes) across all masked positions $m$:
\begin{equation}
\begin{split}
\mathcal{L}_{\text{DR}} = \frac{1}{2} \Bigg(
  \frac{\sum_{i=1}^{N} \sum_{j=1}^{D_{\text{cat}}} m_{i,j}\, \text{CE}(\hat{y}_{i,j},\, y_{i,j})}
       {\sum_{i,j} m_{i,j}} \\
  +\;
  \frac{\sum_{i=1}^{N} \sum_{k=1}^{D_{\text{con}}} m_{i,k}\, \text{KL}(p_{i,k} \parallel q_{i,k})}
       {\sum_{i,k} m_{i,k}}
\Bigg)
\end{split}
\end{equation}

\section{Experiments and Results}
\begin{table}
\caption{Comparison with SOTA methods on SLICE-3D, HOP, and EyePACS datasets. Best results are in \textbf{bold}, second best are \underline{underlined} (Max pAUC = 0.200).}\label{tab:main_results}
\centering
\resizebox{\textwidth}{!}{%
{\fontsize{8}{9}\selectfont
\begin{tabular}{l | cc | cc | cc | cc}
\hline
\multirow{3}{*}{Method} & \multicolumn{2}{c|}{SLICE-3D (ID)} & \multicolumn{2}{c|}{SLICE-3D (OOD)} & \multicolumn{2}{c|}{HOP} & \multicolumn{2}{c}{EyePACS} \\
\cline{2-3}\cline{4-5}\cline{6-7}\cline{8-9}
 & AUC & pAUC & AUC & pAUC & AUC & pAUC & ACC & QWK \\
 & LP/FT & LP/FT & LP/FT & LP/FT & LP/FT & LP/FT & LP/FT & LP/FT \\
\hline
\multicolumn{9}{l}{\textit{Supervised Methods}} \\
\hline
ViT-B\cite{dosovitskiy2020image} & 0.910 & 0.138 & 0.832 & 0.096 & 0.837 & 0.091 & 0.513 & 0.523 \\
CatBoost\cite{prokhorenkova2018catboost} & 0.913 & 0.139 & 0.843 & 0.098 & 0.851 & 0.094 & 0.496 & 0.501 \\
FT-Trans.\cite{gorishniy2021revisiting} & 0.890 & 0.131 & 0.780 & 0.083 & 0.814 & 0.085 & 0.466 & 0.453 \\
DAFT\cite{wolf2022daft} & 0.921 & 0.142 & 0.870 & 0.123 & 0.876 & 0.121 & 0.526 & 0.528 \\
TabPFN v2\cite{hollmann2025accurate} & 0.892 & 0.127 & 0.870 & 0.119 & 0.881 & 0.129 & 0.518 & 0.520 \\
Dino v3\cite{simeoni2025dinov3} & 0.923 & 0.135 & 0.869 & 0.121 & 0.885 & 0.134 & 0.561 & 0.580 \\
\hline
\multicolumn{9}{l}{\textit{SSL Pre-training Methods}} \\
\hline
SimCLR\cite{chen2020simple} & 0.926/0.929 & 0.143/0.152 & 0.843/0.855 & 0.107/0.112 & 0.847/0.858 & 0.116/0.123 & 0.523/0.527 & 0.538/0.540 \\
SCARF\cite{bahri2021scarf} & 0.910/0.920 & 0.137/0.141 & 0.839/0.861 & 0.111/0.109 & 0.829/0.845 & 0.107/0.118 & 0.520/0.524 & 0.530/0.535 \\
SAINT\cite{somepalli2021saint} & 0.930/0.941 & 0.137/0.147 & 0.846/0.869 & 0.115/0.124 & 0.850/0.865 & 0.127/0.129 & 0.567/0.583 & 0.593/0.631 \\
MMCL\cite{hager2023best} & 0.951/0.956 & 0.168/0.170 & 0.871/0.889 & 0.122/0.125 & 0.889/0.871 & 0.128/0.131 & 0.652/0.660 & 0.698/0.704 \\
TIP\cite{du2024tip} & \underline{0.969}/\underline{0.971} & \underline{0.178}/\underline{0.184} & \underline{0.909}/\underline{0.911} & \underline{0.133}/\underline{0.141} & 0.904/\underline{0.912} & 0.131/0.132 & \underline{0.703}/\underline{0.709} & \underline{0.723}/\underline{0.725} \\
CITab\cite{fu2025unleashing} & 0.958/0.964 & 0.171/0.173 & 0.890/0.902 & 0.127/0.130 & \underline{0.906}/0.911 & \underline{0.133}/\underline{0.136} & 0.689/0.695 & 0.720/0.723 \\
\hline
\textbf{AID (Ours)} & \textbf{0.984}/\textbf{0.986} & \textbf{0.192}/\textbf{0.193} & \textbf{0.942}/\textbf{0.944} & \textbf{0.161}/\textbf{0.162} & \textbf{0.919}/\textbf{0.926} & \textbf{0.143}/\textbf{0.147} & \textbf{0.736}/\textbf{0.740} & \textbf{0.753}/\textbf{0.758} \\
\hline
\end{tabular}}}
\end{table}

\begin{table}
\caption{Ablation study on SLICE-3D, HOP, and EyePACS datasets.}\label{tab:main_ablation}
\centering
\resizebox{\textwidth}{!}{%
{\fontsize{8}{9}\selectfont
\begin{tabular}{l | cc | cc | cc | cc}
\hline
\multirow{3}{*}{Ablation Study} & \multicolumn{2}{c|}{SLICE-3D (ID)} & \multicolumn{2}{c|}{SLICE-3D (OOD)} & \multicolumn{2}{c|}{HOP} & \multicolumn{2}{c}{EyePACS} \\
\cline{2-3}\cline{4-5}\cline{6-7}\cline{8-9}
 & AUC & pAUC & AUC & pAUC & AUC & pAUC & ACC & QWK \\
 & LP/FT & LP/FT & LP/FT & LP/FT & LP/FT & LP/FT & LP/FT & LP/FT \\
\hline
w/o SSL pre-training & 0.927/0.927 & 0.143/0.143 & 0.878/0.878 & 0.132/0.132 & 0.848/0.848 & 0.097/0.097 & 0.646/0.646 & 0.632/0.632 \\
w/o DR (ITC+ITM only) & 0.961/0.969 & 0.167/0.172 & 0.933/0.937 & 0.151/0.155 & 0.895/0.906 & 0.132/0.135 & 0.695/0.699 & 0.706/0.713 \\
w/o Adaptive Masking & 0.959/0.966 & 0.168/0.170 & 0.929/0.930 & 0.149/0.149 & 0.891/0.898 & 0.130/0.133 & 0.688/0.697 & 0.689/0.701 \\
w/o Discretization & 0.968/0.973 & 0.172/0.183 & 0.936/0.939 & 0.155/0.157 & 0.904/0.911 & 0.133/\underline{0.141} & 0.704/0.709 & 0.711/0.713 \\
Equal-width Discretization & 0.946/0.950 & 0.163/0.165 & 0.903/0.914 & 0.136/0.142 & 0.887/0.892 & 0.126/0.130 & 0.685/0.684 & 0.696/0.689 \\
Gaussian kernel & \underline{0.976}/\underline{0.978} & \underline{0.187}/\underline{0.186} & \underline{0.938}/\underline{0.940} & \underline{0.158}/\underline{0.159} & 0.908/\underline{0.917} & \underline{0.137}/\underline{0.141} & \underline{0.710}/\underline{0.721} & \underline{0.728}/\underline{0.736} \\
Hard-Label Discretization & 0.972/0.977 & 0.183/0.185 & 0.931/0.939 & 0.151/0.157 & \underline{0.910}/0.916 & 0.136/0.140 & 0.708/0.716 & 0.718/0.731 \\
\hline
\textbf{AID (Ours)} & \textbf{0.984}/\textbf{0.986} & \textbf{0.192}/\textbf{0.193} & \textbf{0.942}/\textbf{0.944} & \textbf{0.161}/\textbf{0.162} & \textbf{0.919}/\textbf{0.926} & \textbf{0.143}/\textbf{0.147} & \textbf{0.736}/\textbf{0.740} & \textbf{0.753}/\textbf{0.758} \\
\hline
\end{tabular}}}
\end{table}

\subsection{Experimental Setup}
The primary dataset, SLICE-3D\cite{kurtansky2024slice}, comprises over 400,000 standardized lesion image tiles from 1,042 patients, positive-to-negative sample ratio 1\textperthousand. To ensure rigorous evaluation, 80,433 samples from 220 patients are geographically isolated as an Out-of-Domain (OOD) test set and strictly excluded from pre-training. The remaining 320,626 images from 822 patients are split into an 85/15 ratio for pre-training and validation. For downstream fine-tuning (FT) and linear probing (LP), 10,228 samples from the 822 patients are reserved as the In-Domain (ID) test set, while the remaining 310,398 samples undergo a 5-fold patient-stratified cross-validation. For downstream evaluation, unimodal and multimodal representations are processed by a three-head ensemble mechanism that averages distinct classification logits to generate final diagnostic predictions. Evaluations are reported on both the ID and OOD test sets. External validation utilizes HOP, a private dataset containing 208,540 images from 284 patients. Furthermore, the EyePACS dataset\cite{diabetic-retinopathy-detection}, comprising 88,702 retinal fundus images processed via the AutoMorph pipeline\cite{zhou2022automorph}, provides a rigorous cross-specialty ophthalmology benchmark. While the UK Biobank dataset\cite{bycroft2018uk} is inaccessible due to restricted paywalled access, the EyePACS dataset is open-source, and we will publicly release its corresponding tabular data. During downstream evaluation, class imbalance is mitigated through weighted sampling. Evaluation metrics include AUROC, Accuracy, Quadratic Weighted Kappa (QWK)\cite{diabetic-retinopathy-detection}, and partial AUC at 80\% sensitivity (pAUC@80\%)\cite{kurtansky2025automated}, prioritizing clinical specificity. The pre-training hyperparameter configuration utilizes a base masking rate $r_{\text{base}} = 0.1$, an importance scaling factor and maximum rate $\alpha = r_{\max} = 0.7$, and 50 quantization bins. All hyperparameter settings were empirically established through extensive validation experiments.

\subsection{Main Results}
As presented in Table \ref{tab:main_results}, all baseline experiments were rigorously reproduced under identical experimental settings. The empirical results demonstrate that the proposed framework establishes a new SOTA. On the critical OOD geographic holdout, the model achieves a full fine-tuning AUC of 0.944 and a pAUC of 0.162, significantly outperforming the 0.911 AUC of the strongest semantic-agnostic baseline, TIP. This substantial margin confirms that the semantic-centric design generates representations highly resilient to geographical distribution shifts. Furthermore, on the ID SLICE-3D evaluation, the linear probe performance reaches an AUC of 0.984, which surpasses the 0.971 full fine-tuning AUC of the TIP baseline. This specific outcome indicates that the multimodal feature space produced by the proposed pre-training strategy possesses exceptional linear separability without requiring deep downstream optimization.

Evaluation on the private HOP dataset confirms that these performance gains remain consistent when tested on completely unseen patient populations, achieving a fine-tuning AUC of 0.926. Applying the framework to the EyePACS dataset further validates the universality of the two-dimensional structural hypothesis. In this challenging ophthalmic domain, the model achieves an accuracy of 0.740 and a QWK of 0.758. This cross-specialty success succinctly demonstrates that modeling feature importance and intra-feature duality provides fundamental benefits for multimodal clinical data integration across diverse medical disciplines.


\subsection{Ablation Studies \& Qualitative Analysis}
The ablation study (Table \ref{tab:main_ablation}) isolates the contribution of each proposed module. Removing adaptive masking decreases the out-of-domain (OOD) AUC from 0.944 to 0.930, confirming that the label-free curriculum guides the model to prioritize diagnostically salient features. Evaluating intra-feature structural modeling provides critical insights. Replacing the soft-label objective with standard continuous regression yields an OOD AUC of 0.939. Applying equal-width discretization degrades performance to 0.914, demonstrating that naive binning creates heavily skewed semantic targets. Hard-label quantile discretization recovers the AUC to 0.939. Notably, substituting the proposed triangle kernel with a computationally heavier Gaussian kernel for soft-label generation achieves an AUC of 0.940. However, the proposed soft-label binning module achieves the peak OOD performance of 0.944, demonstrating that the additional computational overhead of the Gaussian kernel does not translate into diagnostic gains over the simpler triangle kernel. This numerical progression provides evidence that dynamically distributing probability mass across adjacent bins via the simple triangle kernel mathematically preserves the inherent ordinal relationships of clinical measurements.

\begin{figure}[t]
  \centering
  \includegraphics[width=0.91\columnwidth]{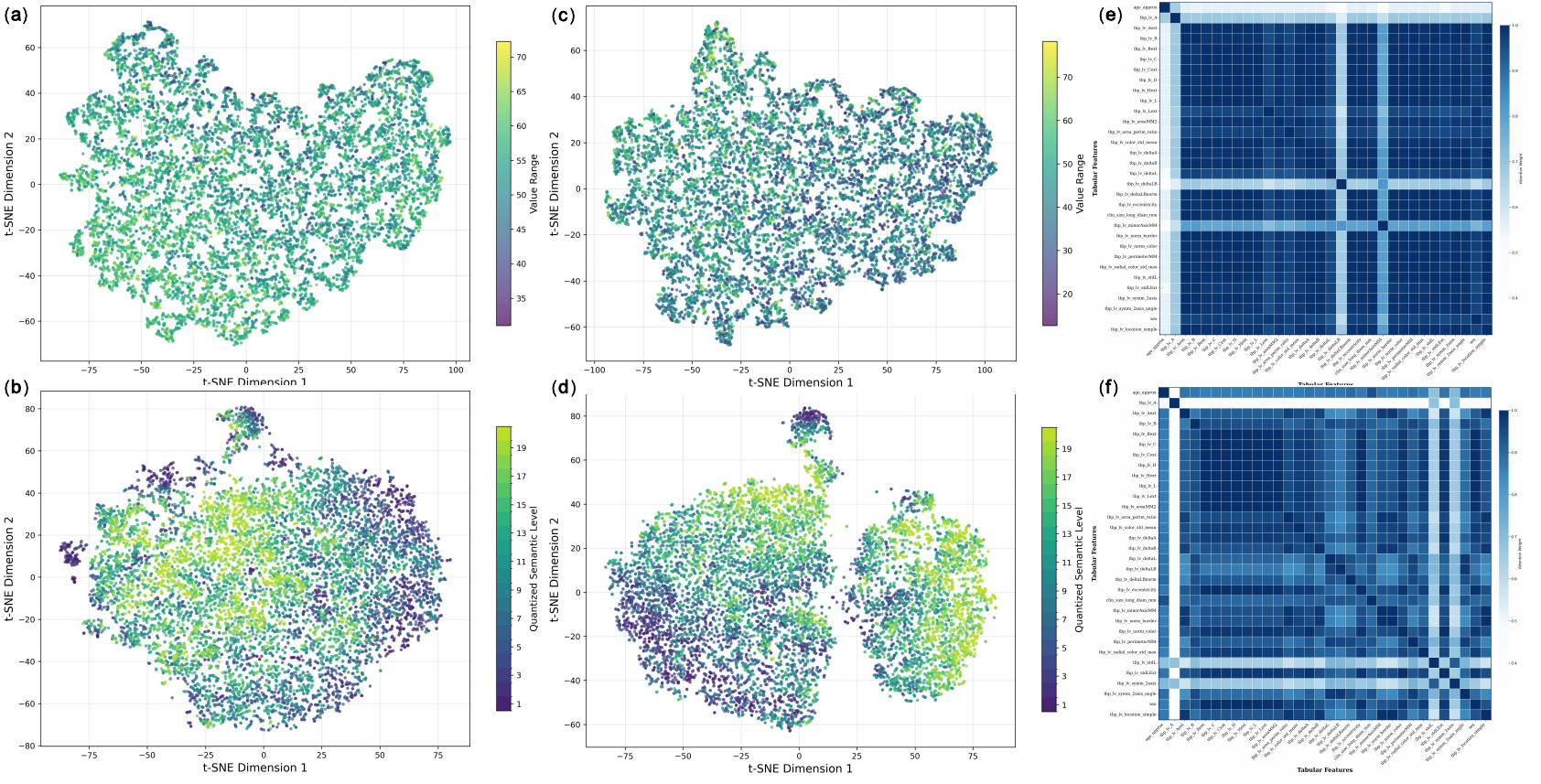}
  \caption{Qualitative comparison between the SOTA baseline (top row) and the proposed framework (bottom row). (a) and (b) visualize t-SNE projections for color feature (tbp\_lv\_H) reconstructions, while (c) and (d) show t-SNE projections for shape feature (tbp\_lv\_L) reconstructions. (e) and (f) present self-attention weight heatmaps from the tabular pathway of the multimodal encoder. }
  \label{fig:tsne}
\end{figure}


As shown in Fig.~\ref{fig:tsne}, visualizations of the learned representations and attention mechanisms corroborate the quantitative improvements. The t-SNE projections contrast the semantic-agnostic baseline against the proposed framework during tabular feature reconstruction. The baseline model maps continuous values into a chaotic, unorganized latent space. Conversely, the soft-label objective successfully transforms this continuous regression space into highly ordered, distinct medical semantic clusters. Furthermore, self-attention heatmaps from the tabular pathway reveal distinct behavioral differences. While the semantic-agnostic baseline exhibits a diffuse, disorganized pan-attention pattern, the proposed model learns a highly structured and importance-aware attention allocation mechanism. This refined attention distribution confirms that the framework efficiently assigns focus based on clinical relevance, faithfully reflecting the hierarchical logic utilized in human diagnostic reasoning.

\section{Conclusion}
We introduce AID, a semantic-aware multimodal pre-training framework that models the inherent two-dimensional structure of medical tabular data. By integrating importance-aware adaptive masking and soft-label discretization, the framework effectively overcomes traditional semantic-agnostic limitations. Evaluations confirm that this framework achieves robust cross-domain generalization and state-of-the-art performance across diverse medical disciplines.

\begin{credits}
\subsubsection{\discintname}
The authors have no competing interests to declare that are relevant to the content of this article.
\end{credits}

\bibliographystyle{splncs04}
\bibliography{reference}

@article{bahri2021scarf,
  title={Scarf: Self-supervised contrastive learning using random feature corruption},
  author={Bahri, Dara and Jiang, Heinrich and Tay, Yi and Metzler, Donald},
  journal={arXiv preprint arXiv:2106.15147},
  year={2021}
}

@inproceedings{chen2020simple,
  title={A simple framework for contrastive learning of visual representations},
  author={Chen, Ting and Kornblith, Simon and Norouzi, Mohammad and Hinton, Geoffrey},
  booktitle={International conference on machine learning},
  pages={1597--1607},
  year={2020},
  organization={PmLR}
}

@inproceedings{du2024tip,
  title={Tip: Tabular-image pre-training for multimodal classification with incomplete data},
  author={Du, Siyi and Zheng, Shaoming and Wang, Yinsong and Bai, Wenjia and O’Regan, Declan P and Qin, Chen},
  booktitle={European Conference on Computer Vision},
  pages={478--496},
  year={2024},
  organization={Springer}
}

@article{hollmann2025accurate,
  title={Accurate predictions on small data with a tabular foundation model},
  author={Hollmann, Noah and M{\"u}ller, Samuel and Purucker, Lennart and Krishnakumar, Arjun and K{\"o}rfer, Max and Hoo, Shi Bin and Schirrmeister, Robin Tibor and Hutter, Frank},
  journal={Nature},
  volume={637},
  number={8045},
  pages={319--326},
  year={2025},
  publisher={Nature Publishing Group UK London}
}

@article{huang2020fusion,
  title={Fusion of medical imaging and electronic health records using deep learning: a systematic review and implementation guidelines},
  author={Huang, Shih-Cheng and Pareek, Anuj and Seyyedi, Saeed and Banerjee, Imon and Lungren, Matthew P},
  journal={NPJ digital medicine},
  volume={3},
  number={1},
  pages={136},
  year={2020},
  publisher={Nature Publishing Group UK London}
}

@article{wolf2022daft,
  title={DAFT: A universal module to interweave tabular data and 3D images in CNNs},
  author={Wolf, Tom Nuno and P{\"o}lsterl, Sebastian and Wachinger, Christian and Alzheimer’s Disease Neuroimaging Initiative and others},
  journal={NeuroImage},
  volume={260},
  pages={119505},
  year={2022},
  publisher={Elsevier}
}

@inproceedings{hager2023best,
  title={Best of both worlds: Multimodal contrastive learning with tabular and imaging data},
  author={Hager, Paul and Menten, Martin and Rueckert, Daniel},
  booktitle={Proceedings of the IEEE/CVF Conference on Computer Vision and Pattern Recognition},
  pages={23924--23935},
  year={2023}
}

@article{kurtansky2024slice,
  title={The SLICE-3D dataset: 400,000 skin lesion image crops extracted from 3D TBP for skin cancer detection},
  author={Kurtansky, Nicholas R and D’Alessandro, Brian M and Gillis, Maura C and Betz-Stablein, Brigid and Cerminara, Sara E and Garcia, Rafael and Girundi, Marcela Alves and Goessinger, Elisabeth Victoria and Gottfrois, Philippe and Guitera, Pascale and others},
  journal={Scientific Data},
  volume={11},
  number={1},
  pages={884},
  year={2024},
  publisher={Nature Publishing Group UK London}
}

@article{dosovitskiy2020image,
  title={An image is worth 16x16 words: Transformers for image recognition at scale},
  author={Dosovitskiy, Alexey and Beyer, Lucas and Kolesnikov, Alexander and Weissenborn, Dirk and Zhai, Xiaohua and Unterthiner, Thomas and Dehghani, Mostafa and Minderer, Matthias and Heigold, Georg and Gelly, Sylvain and others},
  journal={arXiv preprint arXiv:2010.11929},
  year={2020}
}

@article{gorishniy2021revisiting,
  title={Revisiting deep learning models for tabular data},
  author={Gorishniy, Yury and Rubachev, Ivan and Khrulkov, Valentin and Babenko, Artem},
  journal={Advances in neural information processing systems},
  volume={34},
  pages={18932--18943},
  year={2021}
}

@article{prokhorenkova2018catboost,
  title={CatBoost: unbiased boosting with categorical features},
  author={Prokhorenkova, Liudmila and Gusev, Gleb and Vorobev, Aleksandr and Dorogush, Anna Veronika and Gulin, Andrey},
  journal={Advances in neural information processing systems},
  volume={31},
  year={2018}
}

@article{somepalli2021saint,
  title={Saint: Improved neural networks for tabular data via row attention and contrastive pre-training},
  author={Somepalli, Gowthami and Goldblum, Micah and Schwarzschild, Avi and Bruss, C Bayan and Goldstein, Tom},
  journal={arXiv preprint arXiv:2106.01342},
  year={2021}
}

@article{fu2025unleashing,
  title={Unleashing the Power of Image-Tabular Self-Supervised Learning via Breaking Cross-Tabular Barriers},
  author={Fu, Yibing and Zhao, Yunpeng and Zeng, Zhitao and Chen, Cheng and Jin, Yueming},
  journal={arXiv preprint arXiv:2512.14026},
  year={2025}
}

@article{yan2026vision,
  title={A Vision-Language Foundation Model for Zero-shot Clinical Collaboration and Automated Concept Discovery in Dermatology},
  author={Yan, Siyuan and Li, Xieji and Mo, Dan and Tschandl, Philipp and Jiang, Yiwen and Wang, Zhonghua and Hu, Ming and Ju, Lie and Alonso, Cristina and Zheng, Yizhen and others},
  year={2026}
}

@inproceedings{yan2025make,
  title={Make: Multi-aspect knowledge-enhanced vision-language pretraining for zero-shot dermatological assessment},
  author={Yan, Siyuan and Li, Xieji and Hu, Ming and Jiang, Yiwen and Yu, Zhen and Ge, Zongyuan},
  booktitle={International Conference on Medical Image Computing and Computer-Assisted Intervention},
  pages={369--379},
  year={2025},
  organization={Springer}
}

@article{li2025multi,
  title={Multi-Aspect Knowledge-Enhanced Medical Vision-Language Pretraining with Multi-Agent Data Generation},
  author={Li, Xieji and Yan, Siyuan and Liu, Yingsheng and Soyer, H Peter and Janda, Monika and Mar, Victoria and Ge, Zongyuan},
  journal={arXiv preprint arXiv:2512.03445},
  year={2025}
}

@inproceedings{yan2025derm1m,
  title={Derm1m: A million-scale vision-language dataset aligned with clinical ontology knowledge for dermatology},
  author={Yan, Siyuan and Hu, Ming and Jiang, Yiwen and Li, Xieji and Fei, Hao and Tschandl, Philipp and Kittler, Harald and Ge, Zongyuan},
  booktitle={Proceedings of the IEEE/CVF International Conference on Computer Vision},
  pages={12681--12690},
  year={2025}
}

@misc{diabetic-retinopathy-detection,
    author = {Emma Dugas and Jared and Jorge and Will Cukierski},
    title = {Diabetic Retinopathy Detection},
    year = {2015},
    howpublished = {\url{https://kaggle.com/competitions/diabetic-retinopathy-detection}},
    note = {Kaggle}
}

@article{simeoni2025dinov3,
  title={Dinov3},
  author={Sim{\'e}oni, Oriane and Vo, Huy V and Seitzer, Maximilian and Baldassarre, Federico and Oquab, Maxime and Jose, Cijo and Khalidov, Vasil and Szafraniec, Marc and Yi, Seungeun and Ramamonjisoa, Micha{\"e}l and others},
  journal={arXiv preprint arXiv:2508.10104},
  year={2025}
}

@inproceedings{devlin2019bert,
  title={Bert: Pre-training of deep bidirectional transformers for language understanding},
  author={Devlin, Jacob and Chang, Ming-Wei and Lee, Kenton and Toutanova, Kristina},
  booktitle={Proceedings of the 2019 conference of the North American chapter of the association for computational linguistics: human language technologies, volume 1 (long and short papers)},
  pages={4171--4186},
  year={2019}
}

@inproceedings{bhat2021adabins,
  title={Adabins: Depth estimation using adaptive bins},
  author={Bhat, Shariq Farooq and Alhashim, Ibraheem and Wonka, Peter},
  booktitle={Proceedings of the IEEE/CVF conference on computer vision and pattern recognition},
  pages={4009--4018},
  year={2021}
}

@article{kurtansky2025automated,
  title={Automated triage of cancer-suspicious skin lesions with 3D total-body photography},
  author={Kurtansky, Nicholas R and Gillis, Maura C and Codella, Noel CF and D’Alessandro, Brian M and Ge, Zongyuan and Guitera, Pascale and Halpern, Allan C and Kittler, Harald and Malvehy, Josep and Liopyris, Konstantinos and others},
  journal={npj Digital Medicine},
  volume={8},
  number={1},
  pages={708},
  year={2025},
  publisher={Nature Publishing Group UK London}
}

@article{zhou2022automorph,
  title={AutoMorph: automated retinal vascular morphology quantification via a deep learning pipeline},
  author={Zhou, Yukun and Wagner, Siegfried K and Chia, Mark A and Zhao, An and Xu, Moucheng and Struyven, Robbert and Alexander, Daniel C and Keane, Pearse A and others},
  journal={Translational vision science \& technology},
  volume={11},
  number={7},
  pages={12--12},
  year={2022},
  publisher={The Association for Research in Vision and Ophthalmology}
}

@article{bycroft2018uk,
  title={The UK Biobank resource with deep phenotyping and genomic data},
  author={Bycroft, Clare and Freeman, Colin and Petkova, Desislava and Band, Gavin and Elliott, Lloyd T and Sharp, Kevin and Motyer, Allan and Vukcevic, Damjan and Delaneau, Olivier and O’Connell, Jared and others},
  journal={Nature},
  volume={562},
  number={7726},
  pages={203--209},
  year={2018},
  publisher={Nature Publishing Group UK London}
}

@article{lee2020biobert,
  title={BioBERT: a pre-trained biomedical language representation model for biomedical text mining},
  author={Lee, Jinhyuk and Yoon, Wonjin and Kim, Sungdong and Kim, Donghyeon and Kim, Sunkyu and So, Chan Ho and Kang, Jaewoo},
  journal={Bioinformatics},
  volume={36},
  number={4},
  pages={1234--1240},
  year={2020},
  publisher={Oxford University Press}
}

\end{document}